\documentclass[runningheads]{llncs}
\usepackage{graphicx}

\usepackage{tikz}
\usepackage{comment}
\usepackage{amsmath,amssymb} 
\usepackage{color}

\usepackage[accsupp]{axessibility}  

\usepackage{multirow}
\usepackage[colorlinks,linkcolor=red]{hyperref}
\usepackage[misc]{ifsym}

\begin{document}
\pagestyle{headings}
\mainmatter
\def\ECCVSubNumber{3289}  

\title{TS2-Net: Token Shift and Selection Transformer for Text-Video Retrieval} 

\titlerunning{TS2-Net: Token Shift and Selection Transformer for Text-Video Retrieval}
%
\author{Yuqi Liu\inst{1,2}\thanks{This work is done when Yuqi is an intern at Tencent} \and
Pengfei Xiong\inst{2} \and
Luhui Xu\inst{2} \and \\
Shengming Cao\inst{2} \and
Qin Jin\inst{1} \textsuperscript{(\Letter)}
}

\relax\footnotetext{\textsuperscript{\Letter} Corresponding author}
\authorrunning{Y. Liu, P. Xiong, et al.}
%
\institute{School of Information, Renmin University of China \and
Tencent \\
\{yuqi657,qjin\}@ruc.edu.cn, \\
xiongpengfei2019@gmail.com,~\{lukenxu,devancao\}@tencent.com
}

\maketitle

\begin{abstract}
Text-Video retrieval is a task of great practical value and has received increasing attention, among which learning spatial-temporal video representation is one of the research hotspots. The video encoders in the state-of-the-art video retrieval models usually directly adopt the pre-trained vision backbones with the network structure fixed, they therefore can not be further improved to produce the fine-grained spatial-temporal video representation. 
In this paper, we propose Token Shift and Selection Network (TS2-Net), a novel token shift and selection transformer architecture, which dynamically adjusts the token sequence and selects informative tokens in both temporal and spatial dimensions from input video samples. The token shift module temporally shifts the whole token features back-and-forth across adjacent frames, to preserve the complete token representation and capture subtle movements. Then the token selection module selects tokens that contribute most to local spatial semantics. Based on thorough experiments, the proposed TS2-Net achieves state-of-the-art performance on major text-video retrieval benchmarks, including new records on MSRVTT, VATEX, LSMDC, ActivityNet, and DiDeMo. Code will be available at \url{https://github.com/yuqi657/ts2\_net}.

\end{abstract}

\begin{figure*}[t]
    \centering
    \includegraphics[width=1\linewidth]{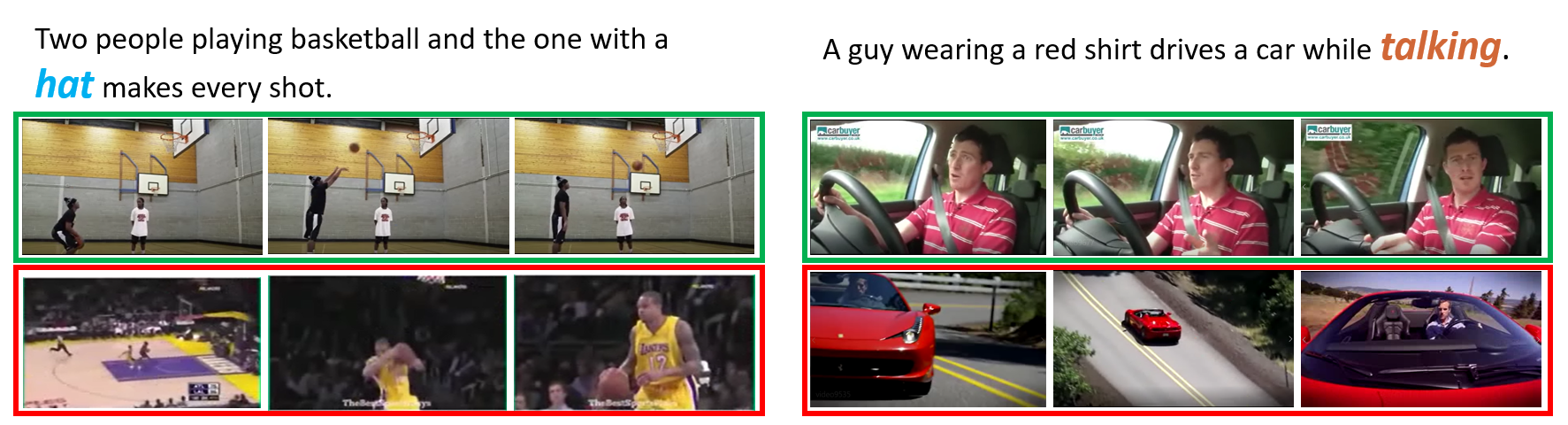}
    \caption{The text-video retrieval examples that require fine-grained video representation. Left: the small object `hat' is important for correctly retrieving the target video. Right: the subtle movement of `talking' is crucial for the correct retrieval of the target video. Green boxes depict the positive video result, while red boxes are negative candidates}
    \label{fig:intro}
\end{figure*}

\section{Introduction}

With advanced digital technologies, massive amount of videos are generated and uploaded online everyday. 
Searching for target videos based on users' text queries is a task of great practical value and has attracted increasing research attention.
Over the past years, different text-video benchmarks  have been established~\cite {anne2017localizing,xu2016msr,chen2011collecting,rohrbach2017movie,wang2019vatex,krishna2017dense} and various text-video retrieval approaches have been proposed~\cite{chen2020fine,dzabraev2021mdmmt,gabeur2020multi,liu2019use,liu2021hit,luo2020univl}, which usually formulate the task as a learning and matching task based on a similarity function between the text query and candidate videos in the corpus.  
With the success of deep neural networks \cite{carreira2017quo,feichtenhofer2019slowfast,xie2018rethinking}, deep learned features have replaced manually-designed features. 
A text-video retrieval engine is generally composed of a text encoder and a video encoder, 
which maps the text query and the video candidate to the same embedding space, where the similarity can be easily computed using a distance metric.

Building a powerful video encoder to produce spatial-temporal feature encoding for videos, that can simultaneously capture motion between video frames, as well as entities in video frames, has been one of the research focuses for text-video retrieval in recent years
\cite{lin2019tsm,arnab2021vivit,liu2021video}.
Lately, Transformer has become the dominant visual encoder architecture, and it enables the training of video-language models with raw video and text data \cite{bain2021frozen,luo2021clip4clip,fang2021clip2video,cheng2021improving}. 
Various video transformers \cite{arnab2021vivit,liu2021video,bertasius2021space,bulat2021space}, considering both spatial and temporal representations, have achieved superior performance on major benchmarks. 
However, these models still lack fine-grained representation capacity in either spatial or temporal dimension. For example, the video encoder in models \cite{luo2021clip4clip,fang2021clip2video,cheng2021improving} normally consists of a single-frame feature extraction module followed by a global feature aggregation module, which lacks fine-grained interaction between adjacent frames and only aggregates the frame-level semantic information. Although the video encoder in Frozen~\cite{bain2021frozen} employs divided space-time attention, it uses only one [CLS] token as the video representation, failing to capture the find-grained spatial-temporal details. In general, all these models can effectively represent obvious motions and categorical spatial semantics in the video, but still lack the capacity for subtle movement and small objects. They will fail in cases such as illustrated in Fig.\ref{fig:intro}, where the video encoder needs to capture the small object (`hat') and subtle movement (`talking') in order to retrieve the correct target videos.  

Based on the structure of video transformer, video sequence is spatially and temporally divided into consecutive patches. To enhance modeling of small objects and subtle movements, patch enhancement is an intuitive and straight-forward solution. This motivates us to find a feasible way to incorporate spatial-temporal patch contexts into encoded features. 
The shift operation is introduced in TSM\cite{lin2019tsm}, which shifts parts of the channel along temporal dimension. Shift Transformer\cite{zhang2021token} applies shift in visual transformer to enhance temporal modeling. However, the architecture of transformer is different from CNN, such partial shift operation damages the completeness of each token representation.

Therefore, in this paper, we propose TS2-Net, a novel token shift and selection transformer network, to realize local patch feature enhancement. Specifically, we first adopt the token shift module in TS2-Net, which shifts the whole spatial token features back-and-forth across adjacent frames, in order to capture local movement between frames. We then design a token selection module to select top-K informative tokens to enhance the salient semantic feature modeling capability.  Our token shift module treats the features of each token as a whole, and iteratively swaps token features at the same location with neighbor frames, to preserve the complete local token representation and capture local temporal semantics at the same time. The token selection module estimates the importance of each token feature of patches with a selection network, which relies on the correlation between all spatial-temporal patch features and [CLS] tokens. It then selects tokens which contributes most to local spatial semantics. Finally, we align cross-modal representation in a fine-grained manner, where we calculate the similarity between text and each frame-wise video embedding and aggregate them together. TS2-Net is optimized with video-language contrastive learning.

We conduct extensive experiments on several text-video retrieval benchmarks to evaluate our model, including MSRVTT, VATEX, LSMDC, ActivityNet, and DiDeMo. Our proposed TS2-Net achieves the state-of-the-art performance on most of the benchmarks. The ablation experiments demonstrate that the proposed token shift and token selection modules both improve the fine-grained text-video retrieval accuracy. The main contributions of this work are as follows:

\begin{itemize}
    \item We propose a new perspective of video-language learning with local patch enhancements to improve the text-video retrieval.
    
    \item We introduce two modules, token shift transformer and token selection transformer, to better model video representation temporally and spatially.
    
    \item We report new records of retrieval accuracy on several text-video retrieval benchmarks. Thorough ablation studies demonstrate the merits of our patch enhancement concept. 
\end{itemize}

\section{Related Work}

\subsection{Video Retrieval}
Various approaches have been proposed to deal with text-video retrieval task, which usually consist of off-line feature extractors and feature fusion module \cite{yu2018joint,liu2019use,gabeur2020multi,dzabraev2021mdmmt,chen2020fine,liu2021hit,croitoru2021teachtext,wang2021t2vlad}. MMT\cite{gabeur2020multi} uses a cross-modal encoder to aggregate feature extracted by different experts. MDMMT\cite{dzabraev2021mdmmt} further utilizes knowledge learned from multi-domain datasets. Recent works \cite{lei2021less,bain2021frozen,luo2021clip4clip,fang2021clip2video,cheng2021improving} attempt to train text-video model in an end-to-end manner. ClipBERT\cite{lei2021less} is the pioneering end-to-end text-video pretrain model. Its promising results show that jointly train high-level semantic alignment network with low-level feature extractor is beneficial. CLIP4Clip\cite{luo2021clip4clip} and CLIP2Video\cite{fang2021clip2video} transfer knowledge from pretrained CLIP\cite{radford2021learning} to video retrieval task. However, these models still lack fine-grained representation capacity in either spatial or temporal dimension. Different from previous works, we aim to model fine-grained spatial and temporal information to enhance text-video retrieval.

\subsection{Visual-Language Pre-training}
Viusal-language pre-training models has shown promising results in visual-and-language tasks such as image retrieval, image caption and video retrieval. 
In works such as Unicoder-VL\cite{li2020unicoder}, VL-BERT\cite{su2019vl} and VLP\cite{zhou2020unified}, text and visual sequence are input into a shared transformer encoder. In Hero\cite{li2020hero}, ClipBERT\cite{lei2021less} and Univl\cite{luo2020univl}, text and visual sequence are encoded independently, then a cross-encoder is used to fuse different modality. While in Frozen\cite{bain2021frozen}, CLIP\cite{radford2021learning}, text and visual sequence are encoded independently and a contrastive loss is used to align text and visual embedding. Our work use the two-stream structure, where text feature and video feature are encoded independently, then a cross-modal contrastive loss is used to align them.

\subsection{Video Representation Learning}
Early works use 2D or 3D-CNN to encode video feature \cite{carreira2017quo,feichtenhofer2019slowfast,feichtenhofer2019slowfast,lin2019tsm}. Recently, Visual Transformer(ViT)\cite{dosovitskiy2020vit} has shown great potential in image modeling. Many works attempt to transfer ViT into video domain~\cite{arnab2021vivit,bertasius2021space,bulat2021space,liu2021video}. TimeSformer\cite{bertasius2021space} and  ViViT\cite{arnab2021vivit} propose variants of spatial-temporal video transformer. There are several works exploring shift operation to enable 2D network learn temporal information, including TSM\cite{lin2019tsm} and Shift Transformer\cite{zhang2021token}. They shift parts of the channel along the temporal dimension. Different from previous work, we consider token shift operation, which we shift all channels of selected visual tokens to the temporal dimension rather than partial shift (i.e. shift some channels).
Token selection has been used to reduce redundancy problem in transformer based visual model. Dynamic ViT\cite{rao2021dynamicvit} and STTS\cite{wang2021efficient} use token selection for efficiency. Perturbed maximum is proposed in \cite{berthet2020learning} to make top-K differentiable. Based on differential top-K\cite{cordonnier2021differentiable}, our work designs a light-weight token selection module to select informative tokens for effective temporal-spatial modeling.

\section{Method}

\begin{figure*}[t]
    \centering
    \includegraphics[width=1\linewidth]{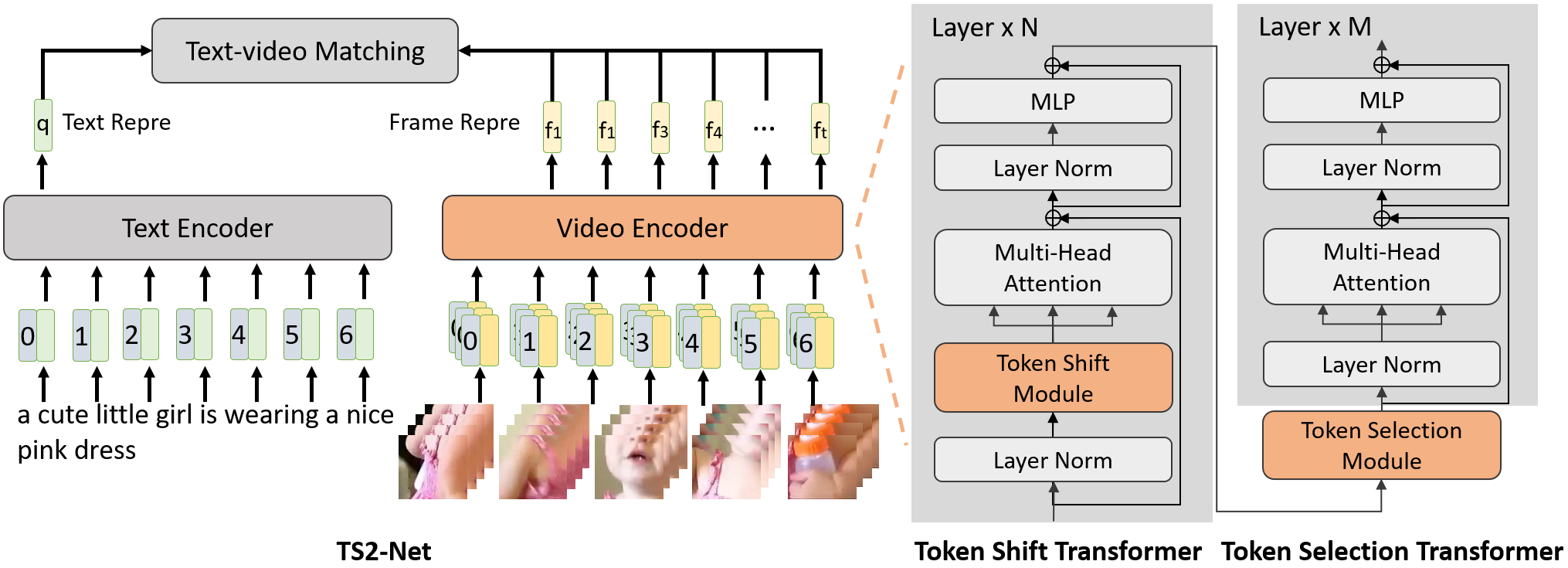}
    \caption{Overview of the proposed TS2-Net model for text-video retrieval, which consists of three key components: the text encoder, the video encoder, and the text-video matching. The video encoder is composed of the Token Shift Transformer and Token Selection Transformer. (`Repre' is short for `Representation')}
    \label{fig:architecture}
\end{figure*}

The goal of text-video retrieval is to find the best matching videos based on the text query. Fig.\ref{fig:architecture} illustrates the overall structure of the proposed TS2-Net model for the text-video retrieval task, which consists of three key components: the text encoder, the video encoder, and the text-video matching. The text encoder encodes the sequence of query words into a query representation $q$. In this paper,
we use GPT~\cite{radford2019language} model as the text encoder. By adding a special token [EOS] at the end of query word sequence, we employ the encoding of [EOS] by the GPT encoder as the query representation $q$. The video encoder encodes the sequence of video frames into a sequence of frame-wise video representation $v=\{f_{1}, f_{2}, \dots, f_{t}\}$. Based on the query and video representation, $q$ and $v$, the text-video matching computes the cross-modal similarity between the query and video candidate.  In following sections, we first elaborate the core ingredients of our video encoder, namely the token shift transformer (Sec.\ref{docu:tokenshift}) and the token selection transformer (Sec.\ref{docu:tokenselect}), and finally present our text-video matching strategy in details (Sec.\ref{docu:textvideomatch}). 



\subsection{Token Shift Transformer}
\label{docu:tokenshift}
\begin{figure*}[t]
    \centering
    \includegraphics[width=1\linewidth]{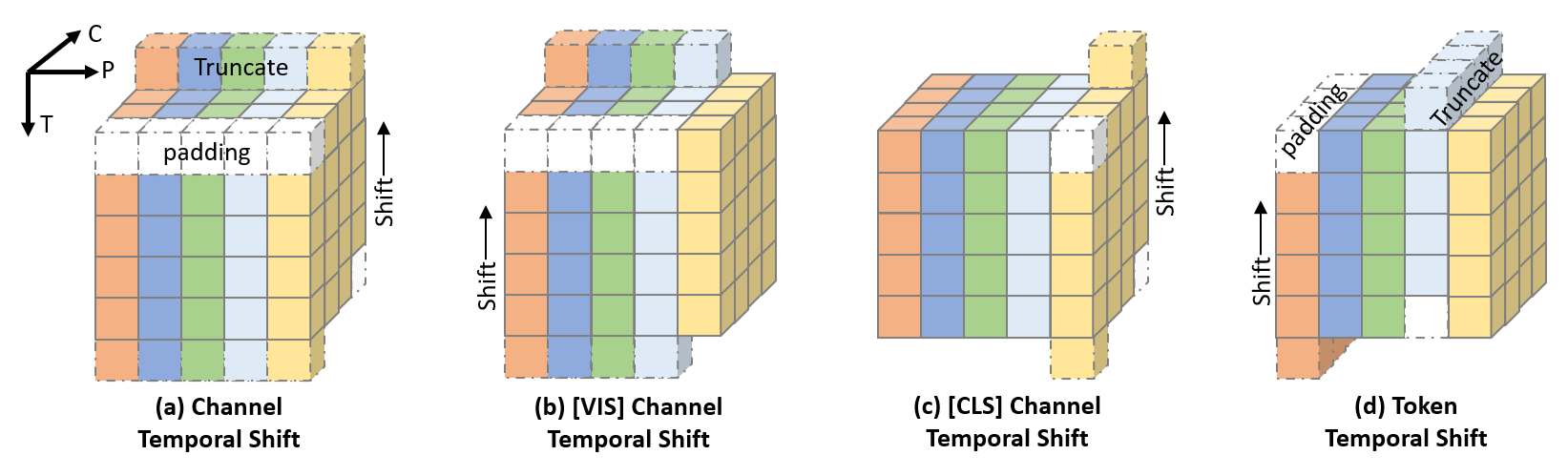}
    \caption{Illustration of different types of Shift operation and our proposed Token Temporal Shift. `T, P, C' refer to video temporal dimension, video token, and feature channel respectively. Each vertical cube group represents a spatial-temporal video token. Cubes with dash line represent tensor truncated, and white cubes represent tensor padding. In Shift-Transformer~\cite{zhang2021token}, tokens are shifted along the channel dimension, while our proposed Token Shift Module does not compromise the integrity of a video token}
    \label{fig:tokenshift}
\end{figure*}

Token shift transformer is based on Vision Transformer (ViT)~\cite{dosovitskiy2020vit}. It inserts a token shift module in the transformer block. Let's review ViT model first, and then describe our modification to ViT. 
Given an image $\boldsymbol{I}$, ViT first splits $\boldsymbol{I}$ into $N$ patches $\{p_{0}, p_{1}, \dots, p_{n-1}\}$. To eliminate ambiguity, we use \textit{token} to represent \textit{patch} below. After adding a [CLS] token $p_{cls}$, the token sequence $\{p_{cls}, p_{0}, p_{1}, \dots, p_{n-1}\}$ is fed into a stack of transformer blocks. Then the image embedding is generated by either averaging all the visual tokens or using the [CLS] token $p_{cls}$. In  this work, we use $p_{cls}$ as the image embedding.
Token shift transformer aims to effectively model subtle movements in a video. The proposed token shift operation is a parameter-free operation, as illustrated in Fig.\ref{fig:tokenshift}. Suppose we have a video $\boldsymbol{V} \in \mathbb{R}^{T \times N \times C}$, where $T$ represents the number of frames, $N$ refers to the number of tokens per frame, and $C$ represents the feature dimension. We feed $T$ frames into ViT to encode frame feature. In certain ViT layer, we shift some tokens from adjacent frames to the current frame to exchange information of adjacent frames. Note that we use a bi-directional token shift in our implementation. By token shift operation across adjacent frames, our model is able to capture subtle movements in the local temporal interval.

Shift-Transformer~\cite{zhang2021token} has also explored several shift variants on the visual transformer architecture. 
Fig.\ref{fig:tokenshift} visualizes the difference between these shift variants and our proposed token shift.  
A naive channel temporal shift swaps part of channels of a frame tensor along temporal dimension, as shown in Fig.~\ref{fig:tokenshift}(a). Shift-Transformer~\cite{zhang2021token} also presents [VIS] channel temporal shift and [CLS] channel temporal shift, as shown in Fig.\ref{fig:tokenshift}(b)(c). They fix tensor in token dimension and shift parts of channels for chosen token along the temporal dimension. Different from these works, our token shift transformer emphasizes the token dimension, where we shift whole channels of a token back-and-forth across adjacent frames, as shown in Fig.\ref{fig:tokenshift}(d). We believe our token shift is better for ViT architecture, because different from the CNN architecture, each token in ViT is independent and contains unique spatial information with respect to its location. Thus shifting parts of channels destroys the integrity of the information contained in a token. On the contrast, shifting a whole token with all channels can preserve complete information contained in a token and enable cross-frame interaction.

However, if we shift most of the tokens in every ViT layer, it damages the spatial modeling ability, and the information contained in these tokens is no longer accessible in the current frame. We therefore use a residual connection between original feature and token shift feature, as illustrated in Fig.\ref{fig:architecture}. In addition, we assume that shallow layers are more important to model spatial features, so shifting in shallow layers could harm spatial modeling. We thus choose to apply token shift operation only in deeper layers in our implementation.

\subsection{Token Selection Transformer}
\label{docu:tokenselect}
Aggregating information from each frame is a necessary step in building the video representation. A naive solution to aggregate per-frame information is by adding some temporal transformer layers, or by mean pooling as CLIP4Clip\cite{luo2021clip4clip}. We argue that aggregation with only the [CLS] token leads to missing important spatial information (i.e. some objects). An alternative way is using all tokens from all frames to aggregate information, 
but this introduces redundancy problem, leading to the pitfall of some background tokens with irrelevant information dominating the final video representation. 

\begin{figure*}[t]
    \centering
    \includegraphics[width=1\linewidth]{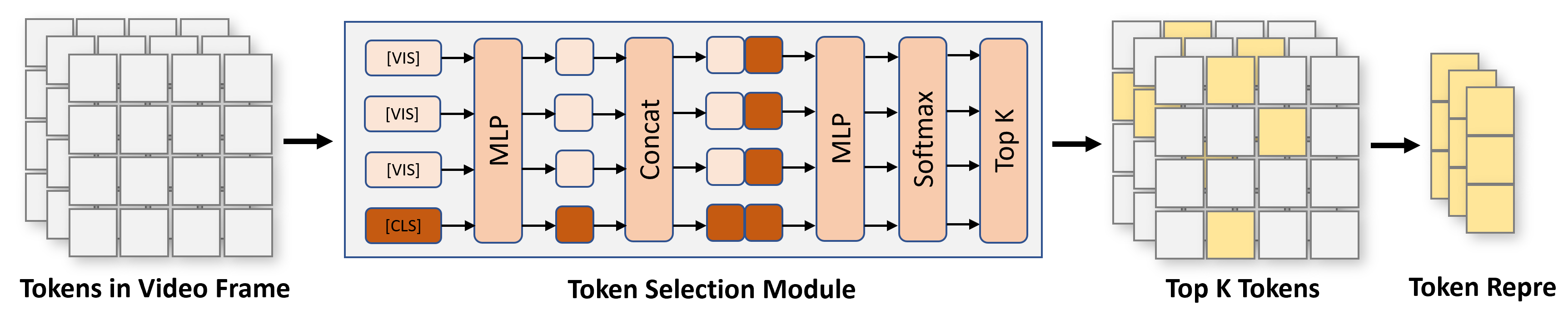}
    \caption{Illustration of Token Selection Module. Top-K informative tokens are selected per frame from original spatial-temporal tokens for following feature aggregation}
    \label{fig:tokenselector}
\end{figure*}

In this work, we propose the token selection transformer by inserting a token selection module, which aims to select informative tokens per frame, especially those tokens containing salient semantics of objects, for video feature aggregation. 
As shown in Fig.\ref{fig:tokenselector}, top-K informative tokens are selected via the trainable token selection module every frame. 
The input of the token selection module is a sequence of tokens of each frame $\boldsymbol{I} = \{p_{cls}, p_{0}, p_{1}, \dots, p_{n-1}\} \in \mathbb{R}^{(N+1) \times C}$. We first apply an MLP over $\boldsymbol{I}$ for channel dimension reduction and output $\boldsymbol{I}^{\prime} = \{p_{cls}^{\prime},p_{0}^{\prime}, p_{1}^{\prime}, \dots, p_{n-1}^{\prime}\} \in \mathbb{R}^{(N+1) \times \frac{C}{2}}$. We then use $p_{cls}^{'}$ as a global frame feature and concatenate it with each local token $p_{i}^{\prime}$,  $\hat{p}_{i}=\left[{p}^{\prime}_{cls}, {p}^{\prime}_{i}\right], 0 \leq i < N$.
We finally feed all the concatenated token features to another MLP followed by a Softmax layer to predict the importance scores, which can be formulated as:

\begin{equation}
\begin{aligned}
\label{eq:topk}
\boldsymbol{S}=\operatorname{Softmax}(\operatorname{MLP}({\hat{p}})) \in \mathbb{R}^{(N+1)}.
\end{aligned}
\end{equation}

\noindent{}We select indices of K most informative tokens based on $\boldsymbol{S}$, denoting as $\mathbf{M}  \in \{0, 1\}^{(N+1) \times K} $, where each column in $\mathbf{M}$ is a one-hot $(N+1)$ dimensional indicator. We extract top-K most informative tokens by:

\begin{equation}
\begin{aligned}
\label{eq:tokenselect}
\mathbf{\hat{I}} = \mathbf{M}^{T} \mathbf{I} ,
\end{aligned}
\end{equation}

\noindent{}After top-K token select on every frame, we input the selected tokens from all frames to a joint spatial-temporal transformer, to learn global spatial-temporal video representation. We also pick the most informative token from each frame as the frame-wise video encoding.

\subsubsection{Differentiable TopK.}
Until now, both top-K operation and one-hot operation are non-differentiable. To make token selection module differentiable, we employ the perturbed maximum method proposed in \cite{berthet2020learning}. Specifically, a discrete optimization problem with input $\boldsymbol{S} \in \mathbb{R}^{(N+1)}$ ($\boldsymbol{S}$ is the importance score matrix in Eq.\ref{eq:topk}) and optimization variable $\mathbf{M} \in \mathbb{R}^{(N+1) \times K}$ ($\mathbf{M}$ is the index indicator matrix in Eq. \ref{eq:tokenselect}) can be formulated as:
\begin{equation}
\label{eq:difftopk}
F(\boldsymbol{S})=\max _{\mathbf{M} \in \mathcal{C}}\langle \mathbf{M} , \boldsymbol{S}\rangle , \mathbf{M}^{*}(\boldsymbol{S})=\underset{\mathbf{M} \in \mathcal{C}}{\arg \max }\langle \mathbf{M} , \boldsymbol{S}\rangle ,
\end{equation}

\noindent{}where $F(\boldsymbol{S})$ represents the top-K selection operation, $\mathbf{M}^{*}(\boldsymbol{S})$ represents the optimal value. Based on Eq.\ref{eq:difftopk}, we can select top-K informative tokens by $F(\boldsymbol{S})$. We calculate forward and backward pass following \cite{abernethy2016perturbation,cordonnier2021differentiable}.

\subsection{Text-Video Matching}
\label{docu:textvideomatch}
The similarity between the text query and video candidate is computed by integrating the similarity between the query and each video frame. 
To be specific, given the query representation $q$ and a sequence of frame-wise video representation $v = \{f_{1}, f_{2}, ...,  f_{t}\}$, we compute the frame-level similarity as follows:

\begin{align}
s_{i}=\frac{q \cdot f_{i}}{\left\|q\right\|\left\|f_{i}\right\|} .
\end{align}

\noindent{}The final text-video matching similarity is defined as the weighted combination of frame-level similarities:

\begin{align}
\label{equation:matchscore}
s=\sum_{i=1}^{n} \alpha_{i} s_{i} ,
\end{align}

\noindent{}where $\alpha_{i}=\frac{\exp \left(\lambda s_{i}\right)}{\sum_{i=1}^{n} \exp \left(\lambda s_{i}\right)}$
and $\lambda$ is a temperature parameter. We set $\lambda$ as 4 empirically in our experiments.
 
Symmetric cross-entropy loss is adopted as our training objective function. For each training step with B text-video pairs, we calculate symmetric cross-entropy loss as follows:

\begin{equation}
\mathcal{L}_{t}^{t 2 v} =-\frac{1}{B} \sum_{i}^{B} \log \frac{\exp \left(\tau \cdot  \operatorname{sim}\left(q_{i}, v_{i}\right)\right)}{\sum_{j=1}^{B} \exp \left(\tau \cdot \operatorname{sim}\left(q_{i}, v_{j}\right)\right)} , 
\end{equation}

\begin{equation}
\mathcal{L}_{t}^{v 2 t} =-\frac{1}{B} \sum_{i}^{B} \log \frac{\exp \left(\tau \cdot \operatorname{sim}\left(q_{i}, v_{i}\right)\right)}{\sum_{j=1}^{B} \exp \left(\tau \cdot \operatorname{sim}\left(q_{j}, v_{i}\right)\right)} ,
\end{equation}

\begin{equation}
\mathcal{L}=\frac{1}{2}\left(\mathcal{L}_{t 2 v}+\mathcal{L}_{v 2 t}\right) ,
\end{equation}

\noindent{}where $\tau$ is a trainable scaling parameter and $\operatorname{sim}\left(q, v\right)$ is calculated using Eq.\ref{equation:matchscore}. During inference, we calculate the matching score between each text and video based on Eq.\ref{equation:matchscore}, and return videos with the highest ranking.

\section{Experiment}

In this section, we carry out text-video retrieval evaluations on multiple benchmark datasets to validate our proposed model TS2-Net. We first ablate the core ingredients of our video encoder, the token shift transformer and the token selection transformer, on the dominant MSR-VTT dataset. We then compare our model with other state-of-the-art models on multiple benchmark datasets 
quantitatively and qualitatively. 


\subsection{Experimental Settings}
\label{docu:setting}

\textbf{Datasets.}
\label{docu:dataset}
To demonstrate the effectiveness and generalization ability of our model, we conduct evaluations on five popular text-video benchmarks, including MSR-VTT\cite{xu2016msr}, VATEX\cite{wang2019vatex}, LSMDC\cite{rohrbach2017movie}, ActivityNet-Caption\cite{caba2015activitynet,krishna2017dense}, DiDeMo\cite{anne2017localizing}. All these datasets are collected from different scenarios with various amounts of captions. Videos in different datasets also have different content styles and different lengths. 

\begin{itemize}
    \item [$\bullet$] \textbf{MSR-VTT}\cite{xu2016msr} contains 10,000 video clips with 20 captions per video. 
    Our experiments follow 1k-A split protocol used in \cite{gabeur2020multi,liu2019use,miech2019howto100m}, where the training set has 9,000 videos with its corresponding captions and test set has 1,000 text-video pairs.
    \item [$\bullet$] \textbf{VATEX}\cite{wang2019vatex} contains 34,991 video clips with several captions per video. We follow HGR\cite{chen2020fine} split protocol. There are 25,991 videos in the training set, 1,500 videos in the validation set and 1,500 videos in the test set.
    \item [$\bullet$] \textbf{LSMDC}\cite{rohrbach2017movie} contains 118,081 video clips, which are extracted from 202 movies. Each video clip has one caption. There are about 100k videos in the training set, 7,408 videos in the validation set and 1,000 videos in the test set. Especially, videos in the test set are from movies disjoint with the training and validation set.
    \item [$\bullet$] \textbf{ActivityNet-Caption}\cite{caba2015activitynet,krishna2017dense} contains 20,000 YouTube videos. Following the same setting as in \cite{luo2021clip4clip,zhang2018cross,gabeur2020multi}, we regard it as a paragraph-video retrieval by concatenate all descriptions of a video. We train our model on \textit{train} split and test our model on \textit{val1} split.
    \item [$\bullet$] \textbf{DiDeMo}\cite{anne2017localizing} contains over 10k videos. There are 8,395 videos in the training set, 1,065 videos in the validation set and 1,004 videos in the test set. Following the same setting as in \cite{luo2021clip4clip,liu2019use,lei2021less}, we concatenate all descriptions of a video to retrieval videos with paragraphs. 
\end{itemize}

\noindent{}\textbf{Evaluation Metrics.} We measure the retrieval performance using standard text-video retrieval metrics: Recall at K (R@K, higher is better), Median Rank (MdR, lower is better) and Mean Rank (MnR, lower is better). R@K calculates the fraction of correct videos among the top K retrieved videos. Similar to previous works \cite{luo2021clip4clip,liu2019use,cheng2021improving}, we use K=1,5,10 for different datasets. We also sum up all the R@K results as $\mathrm{rsum}$ to reflect the overall retrieval performance. 
MedR calculates the median rank of correct results in the retrieval ranking list and MeanR calculates the mean rank of correct results in the retrieval ranking list.

\noindent{}\textbf{Implementation Details.}
\label{docu:imple}
The layer of GPT, token shift transformer and token selection transformer is 12, 12 and 4, respectively. The dimension of text embedding and frame embedding is 512. We initialize transformer layers in GPT, token shift transformer and token selection transformer with pre-trained weight from CLIP(ViT-B/32)\cite{radford2021learning}, using parameters with similar dimension, while other modules are initialized randomly. We choose 4 most informative tokens in MSR-VTT, VATEX, ActivityNet-Caption, DiDeMo, and 1 in LSMDC. We set the max query text length as 32 and max video frame length as 12 in MSR-VTT, VATEX, LSMDC. For ActivityNet-Caption and DiDeMo, we set the max query text length and max video frame length as 64. We train our model with Adam\cite{kingma2014adam} optimizer and adopt a warmup\cite{goyal2017accurate} setting. We choose a batch size of 128. The learning rate of GPT and token shift transformer is 1e-7 and the learning rate of token selection transformer is 1e-4.

\subsection{Ablation Experiments}
\label{docu:ablation}



In this section, we evaluate the proposed token shift transformer and token selection transformer under different settings to validate their effectiveness. We conduct ablation experiments with the 1k-A test split on MSR-VTT\cite{xu2016msr}. We set our baseline model as the degraded TS2-Net model which removes the token shift and token selection modules from TS2-Net.

\begin{table*}[t]
	\begin{center}
	\caption{Performance comparison with different parameter settings of the Token Shift Transformer on MSR-VTT-1k-A test split}
	\label{table:shift}
    \scalebox{0.9}{
	\begin{tabular}{lcc|cccc|cccc|c} 
	\hline
    \multicolumn{3}{c}{} & \multicolumn{4}{c}{Text $\Longrightarrow$ Video} & \multicolumn{4}{c}{Video $\Longrightarrow$ Text} \\
    \hline
	Method & Layers & Ratio & R@1 & R@5 & R@10 & MnR & R@1 & R@5 & R@10 & MnR & rsum \\
    \hline
     Baseline         & - & - & 45.4 & 74.3  & 82.7 & 13.6 & 44.5 & 72.3 & 82.3  & 9.8 & 401.5 \\
     w/ Token Shift & 1-12 &  25\%   & 42.8 & 71.2 & 80.9 & 14.4 & 43.2  & 70.3 & 80.4 & 11.3 & 388.8 \\
     w/ Token Shift & 3-12 &  25\%   & 44.1 & 71.0 & 81.8 & 14.5 & 43.5  & 71.2 & 81.8 & 10.8 & 393.4 \\
     w/ Token Shift & 5-12 &  25\%   & 44.4 & 71.9 & 81.6 & 14.6 & 44.8  & 72.0 & 80.6 & 11.3 & 395.3 \\
     w/ Token Shift & 7-12 &  25\%   & 44.1 & 72.3 & 82.9 & 13.6 & 43.8 & 72.3 & 82.1 & 10.3 & 397.5 \\
     w/ Token Shift & 9-12 &  25\%   & 45.2 & 73.8 & 83.1 & 13.4 & 45.3  & 72.1 & 82.5 & 9.5 & 402 \\
    \hline
     w/ Token Shift & 11-12 &  12.5\%   & 46.0 & 73.3 & 82.2 & 13.8 & \textbf{45.8} & 72.9 & 83.0 & 9.5 & 403.2 \\
     w/ Token Shift & 11-12 &  50\%   & 46.1 & \textbf{74.5} & 83.3 & 13.3 & 45.6 & 72.9 & 82.2 & 9.5 & 404.6 \\
    w/ Token Shift & \textbf{11-12} & \textbf{25\%}  & \textbf{46.2} & 73.9 & \textbf{83.8}  & \textbf{13.0}  & 45.6  & \textbf{73.5}  & \textbf{83.2}  & \textbf{9.3} & \textbf{406.2} \\
    \hline
	\end{tabular} }
	\end{center}
\end{table*}

\begin{table*}[t]
	\begin{center}
	\caption{Performance comparison between other shift operation variants and our proposed token shift module on MSR-VTT-1k-A test split}
	\label{table:shift2}
    \scalebox{0.9}{
	\begin{tabular}{lc|cccc|cccc|c} 
	\hline
    \multicolumn{2}{c}{} & \multicolumn{4}{c}{Text $\Longrightarrow$ Video} & \multicolumn{4}{c}{Video $\Longrightarrow$ Text} \\
    \hline
	Method & & R@1 & R@5 & R@10 & MnR & R@1 & R@5 & R@10 & MnR & rsum \\
    \hline
    Baseline            & & 45.4 & 74.3 & 82.7 & 13.6 & 44.5 & 72.3 & 82.3 & 9.8 & 401.5 \\
    Channel Shift\cite{zhang2021token}   & & 45.6 & 73.6 & 83.1 & 13.7 & 45.0 & 73.2 & 82.7 & 9.7 & 403.2 \\
    $[$VIS$]$ Channel Shift\cite{zhang2021token} & & 45.1 & 73.8 & 83.5 & 13.9 & 44.7 & 73.3 & 82.2 & 9.8 & 402.6 \\
    $[$CLS$]$ Channel Shift\cite{zhang2021token} & & 45.8 & \textbf{74.3} & 83.0 & 13.6 & 44.7 & 72.9 & 82.5 & 9.8 & 403.2 \\
    \hline
    \textbf{Token Shift} & & \textbf{46.2} & 73.9 & \textbf{83.8}  & \textbf{13.0}  &    \textbf{45.6}  & \textbf{73.5}  & \textbf{83.2}  & \textbf{9.3} & \textbf{406.2} \\
    \hline
	\end{tabular} }
	\end{center}
\end{table*}

\noindent{}\textbf{Ablation of Token Shift Transformer.} 
We ﬁrst analyze the impact of some factors on the token shift module in Tab.\ref{table:shift}, including shift layer and shift ratio.  Shift layer (in which layers should we insert token shift) and shift ratio (how many tokens should we shift) are two main factors that affect the final retrieval performance. The backbone of our token shift transformer is the 12-layer ViT. We thus experiment to insert the token shift module in different layers. As shown in Tab.\ref{table:shift}, shift operation in deeper layers (i.e. 11-12 layers) brings retrieval performance improvement. 
But if we shift more layers (i.e. 9-12 layers), it hurts the retrieval performance, and it hurts more if we operate shift in shallower layers (i.e. 1-12 layers). 
Our explanation is that shallow layers in ViT are more important in modeling spatial information, so shift in shallow layers damages spatial modeling ability.
We thus choose to insert the token shift module in the 11-12 layers in the following experiments. 
In terms of shift ratio, we find that shifting 25\% tokens back-and-forth across frames achieves the best retrieval performance. Despite some slight fluctuations, token shift with different ratios achieve better results than the baseline model. The improvement is more obvious especially for R@1. 

\begin{figure*}[t]
    \centering
    \includegraphics[width=1\linewidth]{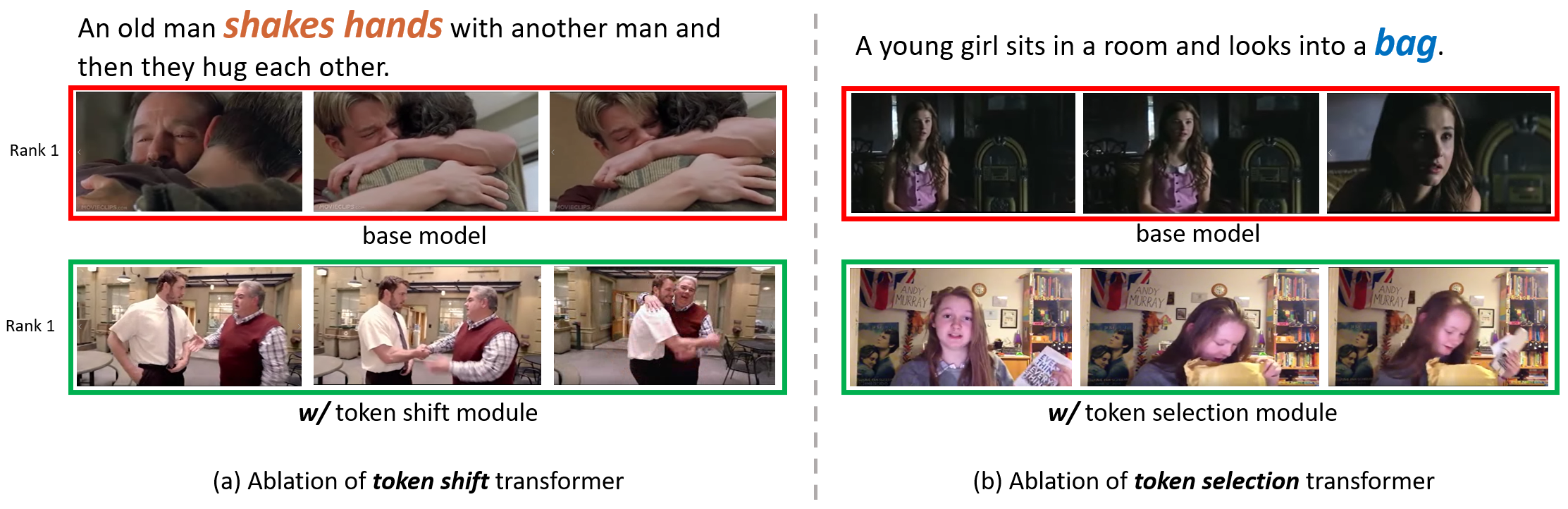}
    \caption{The text-video retrieval results of different network architecture. Left: with \textit{token shift transformer}, our model is able to distinguish `shake hands', while the baseline model retrieves an incorrect video. Right: with \textit{token selection transformer}, our model retrieves the correct video, although `bag' is only shown in small part of video frames. Green boxes: correct target video; red boxes: incorrect target video}
    \label{fig:ablation}
\end{figure*}

We further conduct experiments to compare our proposed token shift module with other shift operation variants proposed in Shift-ViT\cite{zhang2021token}. As shown in Tab.\ref{table:shift2}, our proposed token shift module outperforms all other shift operation variants. This is because our token shift operation can preserve the integrity of the token feature, posing minor impact on the spatial modeling ability. We visualize the retrieval results from the baseline model and the model with token shift transformer in Fig.\ref{fig:ablation}(a). With token shift transformer, the model is able to capture subtle movement such as `shake hand'.

\begin{table*}[ht]
	\begin{center}
	\caption{Comparison results with different settings of Token Selection Transformer}
	\label{table:select}
    \scalebox{0.9}{
	\begin{tabular}{lc|cccc|cccc|c} 
	\hline
    \multicolumn{2}{c}{} & \multicolumn{4}{c}{Text $\Longrightarrow$ Video} & \multicolumn{4}{c}{Video $\Longrightarrow$ Text} \\
    \hline
	Method & top-K & R@1 & R@5 & R@10 & MnR & R@1 & R@5 & R@10 & MnR & rsum \\
    \hline
    Token Shift & 1 & 46.2 & 73.9 & 83.8 & 13.0 & 45.6 & 73.5 & 83.2 & 9.3 & 406.2 \\
    \hline
    w/ all token  & 50 & 45.8 & 73.5 & 83.4 & 13.5 & 44.7 & 73.1 & 82.4 & 9.4 & 402.9 \\
    w/ Random select & 4 & 46.4 & 73.9 & 83.5 & 13.1 & 45.1 & 73.5 & 82.1 & 9.5 & 404.5 \\
    \hline
    w/ Select token & 2 & 47.0 & 74.2 & 83.6 & 13.1 & \textbf{45.6} & 74.0 & 83.5 & 9.3 & 407.9 \\
    w/ Select token & 6 & 46.6 & 74.4 & \textbf{84.3} & 13.2 & 44.5 & 73.8 & 83.2 & 9.2 & 406.8  \\
    w/ Select token & 8 & 46.4 & 73.9 & 83.5 & 13.2 & 45.0 & 74.1 & \textbf{83.9} & 9.2 & 406.8  \\
    \hline
    \textbf{TS2-Net} & 4 & \textbf{47.0}  & \textbf{74.5}  & 83.8  & \textbf{13.0}  & 45.3  & \textbf{74.1}  & 83.7  & \textbf{9.2} & \textbf{408.4} \\
	\hline
	\end{tabular} }
	\end{center}
\end{table*}

\noindent{}\textbf{Ablation of Token Selection Transformer.}
The token selection transformer follows the token shift transformer to select the most informative tokens for the next transformer propagation. We conduct experiment to verify what proportion of tokens is beneficial to the final retrieval in Tab.\ref{table:select}. As can be observed, selecting fewer tokens per frame tends to achieve better performance than selecting more. For example, the R@1 performance decreases from 47.0 to 45.8 while the number of selected tokens increases from 2 to 50.
We consider that fewer informative tokens are sufficient to preserve the salient spatial information, while adding more tokens may bring redundancy problem. Although  random selection also improves the performance slightly, it can not beat the proposed learnable token selection module.
In Fig.\ref{fig:ablation}(b), we show a retrieval case from the baseline model and the model with token selection transformer. With token selection transformer, the model is able to capture the small object `bag' in video frames.

\begin{table*}[t]
	\begin{center}
	\caption{Retrieval results on MSR-VTT-1kA. Other SOTA methods are adopted as comparisons. Note that CLIP2TV uses patch size of 16$\times$16, so we use TS2-Net(ViT16) for fair comparison. All results in this table do not use inverted softmax}
	\label{table:msrvtt}
    \scalebox{0.9}{
	\begin{tabular}{c|ccccc|ccccc} 
	\hline
    \multicolumn{1}{c}{} & \multicolumn{5}{c}{Text $\Longrightarrow$ Video} & \multicolumn{5}{c}{Video $\Longrightarrow$ Text} \\
    \hline
	Method & R@1 & R@5 & R@10 & MdR & MnR & R@1 & R@5 & R@10 & MdR & MnR \\
    \hline
    CE\cite{liu2019use}                    & 20.9 & 48.8 & 62.4 & 6.0 & 28.2 & 20.6 & 50.3 & 64.0 & 5.3 & 25.1\\
    TACo\cite{yang2021taco}                & 26.7 & 54.5 & 68.2 & 4.0 & - & - & - & - & - & - \\
    MMT\cite{gabeur2020multi}              & 26.6 & 57.1 & 69.6 & 4.0  & 24.0 & 27.0 & 57.5 & 69.7 & 3.7 & 21.3 \\
    SUPPORT-SET\cite{patrick2020support}   & 27.4 & 56.3 & 67.7 & 3.0  & -    & 26.6 & 55.1 & 67.5 & 3.0 & - \\
    TT-CE\cite{croitoru2021teachtext}      & 29.6 & 61.6 & 74.2 & 3.0  & - & - & - & - & - & - \\
    T2VLAD\cite{wang2021t2vlad}            & 29.5 & 59.0 & 70.1 & 4.0  & -   & 31.8 & 60.0 & 71.1 & 3.0 & -\\
    HIT-pretrained\cite{liu2021hit}    & 30.7 & 60.9 & 73.2 & 2.6  & -    & 32.1 & 62.7 & 74.1 & 3.0 & - \\
    Frozen\cite{bain2021frozen}            & 31.0 & 59.5 & 70.5 & 3.0 & - & - & - & - & - & - \\
    MDMMT\cite{dzabraev2021mdmmt}          & 38.9 & 69.0 & 79.7	& 2.0  & 16.5 & -    &    - & -    &   - & - \\
    CLIP\cite{radford2021learning}             & 39.7 & 72.3 & 82.2    & 2.0    & 12.8 & 11.3 & 22.7 & 29.2 &  5.0  & - \\
    CLIP4Clip\cite{luo2021clip4clip}           & 44.5 & 71.4 & 81.6 & 2.0 & 15.3 & 42.7 & 70.9 & 80.6 & 2.0 & 11.6\\
    CAMoE\cite{cheng2021improving}             & 44.6 & 72.6 & 81.8 & 2.0 & 13.3 & 45.1 & 72.4 & 83.1 & 2.0 & 10.0\\
    CLIP2Video\cite{fang2021clip2video}       & 45.6 & 72.6 & 81.7 & 2.0 & 14.6 & 43.5 & 72.3 & 82.1 & 2.0 & 10.2\\
    \textbf{TS2-Net} & {\bf 47.0} & {\bf 74.5} & {\bf 83.8} & {\bf 2.0} & {\bf 13.0} & {\bf 45.3} & {\bf 74.1} & {\bf 83.7} & {\bf 2.0} & {\bf 9.2}\\
    \hline
    CLIP2TV\cite{gao2021clip2tv} & 48.3 & 74.6 & 82.8 & 2.0 & 14.9 & 46.5 & 75.4 & 84.9 & 2.0 & 10.2\\
    \textbf{TS2-Net(ViT16)} & {\bf 49.4} & {\bf 75.6} &  {\bf 85.3} & {\bf 2.0} & {\bf 13.5} & {\bf 46.6}  & {\bf 75.9} &  {\bf 84.9} & {\bf 2.0} & {\bf 8.9}\\
	\hline
	\end{tabular}}
	\end{center}
\end{table*}

\subsection{Comparisons with State-of-the-art Models}
\label{docu:sota}

\noindent{}\textbf{MSR-VTT-1kA.} We compare our proposed TS2-Net with other state-of-the-art methods on five benchmarks. Tab.\ref{table:msrvtt} presents the results on MSR-VTT-1kA test set. Our model outperforms previous methods across different evaluation metrics. 
With token shift transformer and token selection transformer, our model is able to capture subtle motion and salient objects, and thus our final video representation contains rich semantics. Compared with video-to-text retrieval, the gain on text-to-video retrieval is more significant. We consider it is because the proposed token shift and token selection modules enhance the video encoder,  while a relative simple text encoder is adopted.

\noindent{}\textbf{Other Benchmarks.} Tab.\ref{table:vatexlsmdc} presents text-to-video retrieval results on VATEX, LSMDC, ActivityNet-Caption and DiDeMo. Results on these datasets demonstrate the generalization and robustness of our proposed model. We can observe that our model achieves consistent improvements across different datasets, which demonstrates that  it is beneficial to encode spatial and temporal features simultaneously by our token shift and token selection. Note that our performance surpasses QB-Norm\cite{bogolin2021cross} on LSMDC and VATEX even without inverted softmax, as shown in Tab.\ref{table:vatexlsmdc}. More detailed analysis will be provided in supplementary materials.

\begin{table*}[t]
	\begin{center}
	\caption{Text-to-Video retrieval results on VATEX, LSMDC, ActivityNet and DiDeMo. QB-Norm uses dynamic inverted softmax during inference, while other methods report results without inverted softmax}
	\label{table:vatexlsmdc}
	\scalebox{0.85}{
	\begin{minipage}{0.62\linewidth}
        \begin{tabular}{c|ccccc} 
        	\hline
            \multicolumn{6}{c}{VATEX} \\
            \hline
        	Method & R@1 & R@5 & R@10 & MdR & MeanR \\
            \hline
        	Dual Enc.\cite{dong2019dual}                 & 31.1 & 67.5 & 78.9    & 3.0    & - \\
        	HGR\cite{chen2020fine}                     & 35.1 & 73.5 & 83.5    & 2.0    & - \\
        	CLIP\cite{radford2021learning}             & 39.7 & 72.3 & 82.2    & 2.0    & 12.8\\
            CLIP4Clip\cite{luo2021clip4clip}             & 55.9 & 89.2 & 95.0 & 1.0 & 3.9 \\
            QB-Norm*\cite{bogolin2021cross}               & 58.8 & 88.3 & 93.8 & 1.0 & - \\
            CLIP2Video\cite{fang2021clip2video}          & 57.3 & 90.0 & \textbf{95.5} & 1.0 & 3.6 \\
            \hline
            \textbf{TS2-Net} & \textbf{59.1} & \textbf{90.0} & 95.2 & \textbf{1.0}  & \textbf{3.5}\\
        	\hline
        	
        	\multicolumn{6}{c}{}\\
        	
        	\hline
            \multicolumn{6}{c}{ActivityNet} \\
            \hline
        	Method & R@1 & R@5 & R@10 & MdR & MeanR \\
            \hline
            CE\cite{liu2019use}                           & 20.5 & 47.7 & 63.9 & 6.0 & 23.1 \\
            ClipBERT\cite{lei2021less}                    & 21.3 & 49.0 & 63.5 & 6.0  & - \\
            MMT-Pretrained\cite{gabeur2020multi}            & 28.7 & 61.4 & - & 3.3  & 16.0 \\
            CLIP4Clip\cite{luo2021clip4clip}             & 40.5 & 73.4 & - & 2.0 & \textbf{7.5} \\
            \hline
            \textbf{TS2-Net} & {\bf 41.0} & {\bf 73.6} &  {\bf 84.5} & \textbf{2.0} & 8.4\\
        	\hline
        	
        \end{tabular}
	\end{minipage}

	\begin{minipage}{0.58\linewidth}
        \begin{tabular}{c|ccccc} 
        	\hline
            \multicolumn{6}{c}{LSMDC} \\
            \hline
        	Method & R@1 & R@5 & R@10 & MdR & MeanR\\
            \hline
             JSFusion\cite{yu2018joint}                           & 9.1 & 21.2 & 34.1 & 36.0 & - \\
             CE\cite{liu2019use}                                  & 11.2 & 26.9 & 34.9 & 25.3 & - \\
             Frozen\cite{bain2021frozen}                 & 15.0 & 30.8 & 39.8 & 20.0 & -\\
             CLIP4Clip\cite{luo2021clip4clip}             & 22.6 & 41.0 & 49.1 & 11.0 & 61.0 \\
             QB-Norm*\cite{bogolin2021cross}               & 22.4 & 40.1 & 49.5 & 11.0 & -\\
             CAMoE\cite{cheng2021improving}               & 22.5 & \textbf{42.6} & 50.9 & - & \textbf{56.5}\\
             
             \hline
            \textbf{TS2-Net} &  {\bf 23.4}  & 42.3 &  {\bf 50.9} & \textbf{9.0} & 56.9\\
        	\hline
        	
        	\multicolumn{6}{c}{}\\
        	
        	\hline
            \multicolumn{6}{c}{DiDeMo} \\
            \hline
        	Method & R@1 & R@5 & R@10 & MdR & MeanR\\
            \hline
            ClipBERT\cite{lei2021less}                    & 20.4 & 48.0 & 60.8 & 6.0  & - \\
            TT-CE\cite{croitoru2021teachtext}           & 21.1 & 47.3 & 61.1 & 6.3 & - \\
            Frozen\cite{bain2021frozen}                 & 31.0 & 59.8 & 72.4 & 3.0 & -\\
            CLIP4Clip\cite{luo2021clip4clip}            & \textbf{42.5} & 70.2 & 80.6 & 2.0 & 17.5\\
            \hline
            \textbf{TS2-Net}  &       41.8 & {\bf 71.6} &  {\bf 82.0} & \textbf{2.0} & \textbf{14.8}\\
        	\hline
        	
        \end{tabular}
	\end{minipage}}
	\end{center}
\end{table*}

\subsection{Qualitative Results}
\label{docu:quatitive}
\begin{figure*}[t]
    \centering
    \includegraphics[width=1\linewidth]{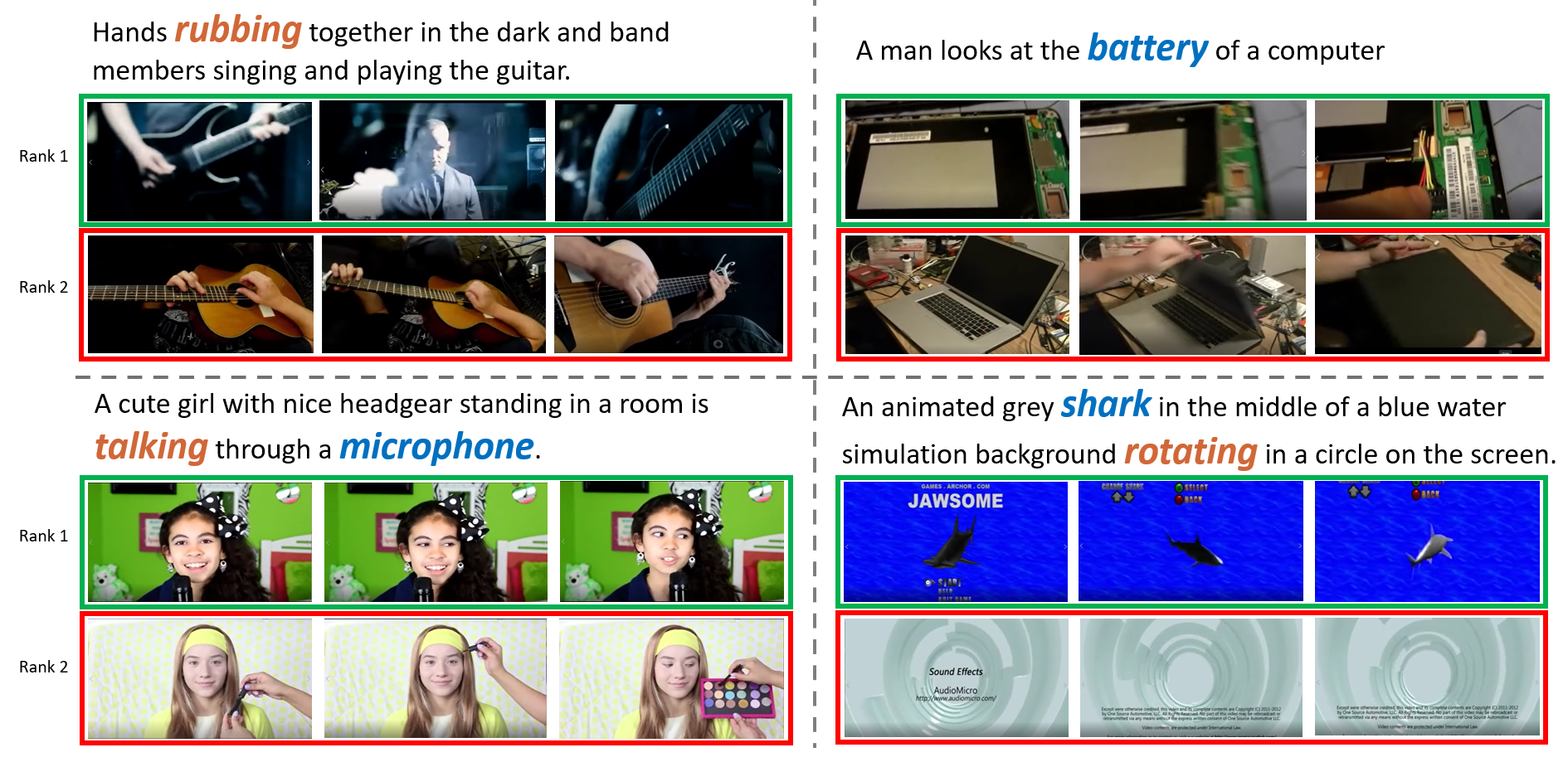}
    \caption{Visualization of text-video retrieval examples. We sorted results based on its similarity scores. Green: ground truth; Red: incorrect}
    \label{fig:qulitative}
\end{figure*}

We visualize some retrieval examples from the MSR-VTT testing set for text-to-video retrieval in Fig.\ref{fig:qulitative}. In the top left example, our model is able to distinguish `hand rubbing' (in the middle picture) during a guitar-playing scene. The bottom right example shows our model can distinguish `computer battery' from `computer'. In the bottom left example, our model retrieves the correct video which contains all actions and objects expressed in the text query, especially the small object `microphone' and tiny movement `talking'. In the bottom right example, our model retrievals the correct result although `rotating' is a periodic movement and is hard to spot. 

We also select a subset from the MSR-VTT-1kA test set. Queries in this subset are selected based on their corresponding video's visual appearance, where objects mentioned in query are shown in a small part of video and movements mentioned in query is slight. Such as `\textit{little pet shop cat} getting a bath and washed with \textit{little brush}', `man \textit{talks} in front of a green bicycle', `dog is drinking milk with \textit{baby nibble} bottle', `a golf player is trying to hit the ball into the \textit{pit}'. Since such cases account for a small proportion, so the total number of this subset is 103. During inference, we calculate similarity between queries in subset with videos in whole test set. We compare our model with another strong baseline on this subset. Our model achieves 79.6 on R@1 metric, while CLIP4Clip\cite{luo2021clip4clip} only achieves 39.8. There is a significant margin and this verifies the effectiveness of TS2-Net in handling local subtle movements and local small entities.

\section{Conclusion}

In this work, we propose Token Shift and Selection Network (TS2-Net), a noval transformer architecture with token shift and selection modules, which aims to further improve the video encoder for better video representation. A token shift transformer is used to capture subtle movements, followed by a token selection transformer to enhance salient object modeling ability. Superior experimental results show our proposed TS2-Net outperforms start-of-the-art methods on five text-video retrieval benchmarks, including MSR-VTT, VATEX, LSMDC, ActivityNet-Caption and DiDeMo.

\noindent{}\textbf{Acknowledgement.} This work was partially supported by National Key R\&D Program of China (No. 2020AAA0108600) and National Natural Science Foundation of China (No. 62072462).

\par\vfill\par
\clearpage

%
%
\bibliographystyle{splncs04}
\bibliography{egbib}

\par\vfill\par
\clearpage
\appendix
\section{Inverted Softmax.} 
\label{docu:dis}
The hubness phenomenon\cite{liu-ye-2019-strong} is that a data point occurs among the k nearest neighbors of other data points. Dual softmax loss (DSL) was mentioned in CAMoE\cite{cheng2021improving}, which adopts a inverted softmax\cite{smith2017offline}.  
QB-Norm\cite{bogolin2021cross} proposes a querybank normalization with dynamic inverted softmax (DIS) to deal with hubness problem. 
CLIP2TV\cite{gao2021clip2tv} also reports its results with inverted softmax. 
We compare their results with basic inverted softmax during inference in Tab.\ref{table:invertedsoftmax}. Our results again surpass all other methods with significant improvement.

\begin{table*}[t]
	\begin{center}
	\caption{Text-to-Video retrieval results with Inverted Softmax}
	\label{table:invertedsoftmax}

	\scalebox{0.85}{
	\begin{minipage}[t]{0.60\linewidth}
        \begin{tabular}{c|ccccc} 
        	\hline
            \multicolumn{6}{c}{MSRVTT-1kA} \\
            \hline
        	Method & R@1 & R@5 & R@10 & MdR & MeanR \\
            \hline
             QB-Norm\cite{bogolin2021cross}               & 47.2 & 73.0 & 83.0 & 2.0 & - \\
             CAMoE\cite{cheng2021improving}               & 47.3 & 74.2 & 84.5 & 2.0 & 11.9 \\
             \textbf{TS2-Net}                              & {\bf 51.1} & {\bf 76.9} &  {\bf 85.6} & {\bf 1.0} & {\bf 11.7}\\
             \hline
        	 CLIP2TV\cite{gao2021clip2tv}                 & 52.9 & 78.5 & 86.5 & 1.0 & 12.8 \\
             \textbf{TS2-Net(ViT16)}                             & {\bf 54.0} & {\bf 79.3} &  {\bf 87.4} & {\bf 1.0} & {\bf 11.7}\\
        	\hline
        	
        \end{tabular}
	\end{minipage}

	\begin{minipage}[ht]{0.56\linewidth}
        \begin{tabular}{c|ccccc} 
        	\hline
        	\multicolumn{6}{c}{DiDeMo} \\
            \hline
        	Method & R@1 & R@5 & R@10 & MdR & MeanR \\
            \hline
             QB-Norm\cite{bogolin2021cross}               & 43.5 & 71.4 & 80.9 & 2.0 & -\\
             CAMoE\cite{cheng2021improving}               & 43.8 & 71.4 & 79.9 & 2.0 & 16.3\\
             \textbf{TS2-Net}                 & \textbf{47.4} & \textbf{74.1} & \textbf{82.4} & \textbf{2.0} & \textbf{12.9}\\
        	\hline
        	\multicolumn{6}{c}{}\\
        	\multicolumn{6}{c}{}\\
        	
        \end{tabular}
	\end{minipage}}
	\end{center}

\end{table*}

\section{Evaluation Summary on Different Benchmarks}

\begin{figure*}[ht]
    \centering
    \includegraphics[width=1\linewidth]{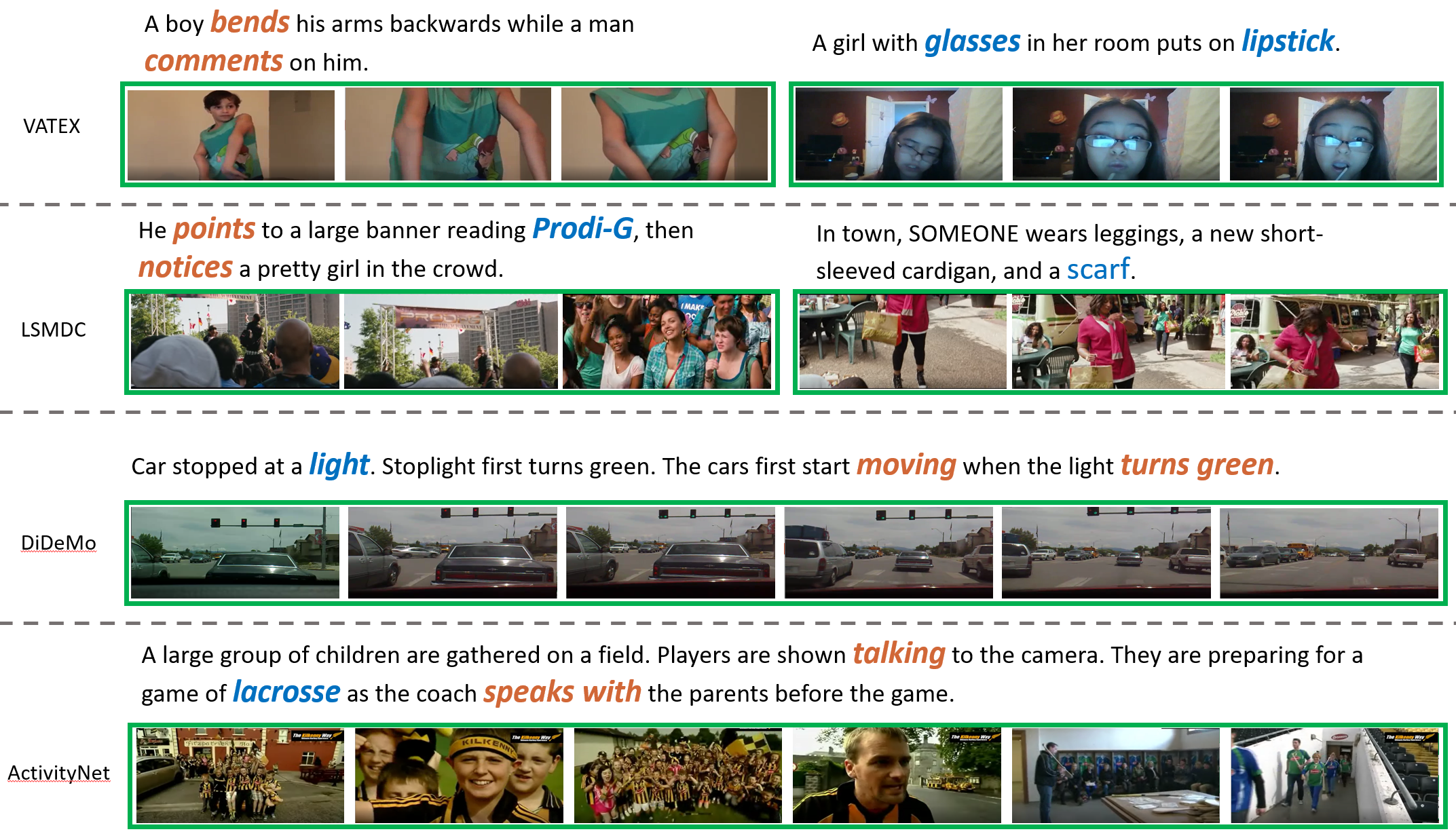}
    \caption{Examples of Text-Video retrieval pairs from different benchmark datasets}
    \label{fig:differentdatacases}

\end{figure*}

We compared our model to other state-of-the-art methods on different video retrieval benchmark datasets in the main paper. Table \ref{table:summary} summarizes the performance comparison between our proposed model TS2-Net and the previous best model on five different benchmark datasets. 
Among all these datasets, VATEX\cite{wang2019vatex} and MSR-VTT\cite{xu2016msr} contain standard captions with average length of 15 words and 8 words respectively. LSMDC\cite{rohrbach2017movie} contains videos in the movie domain. There is no movie overlap between the training and test set. So it can verify the generalization ability of a model. 
ActivityNet-Caption\cite{caba2015activitynet,krishna2017dense} and DiDeMo\cite{anne2017localizing} offer paragraph-video retrieval, which means that the query text involves multi sentences and the video duration is long. Therefore, the query text in these two datasets contains more semantic information. 
We show some examples of these datasets in Fig.\ref{fig:differentdatacases}. 
Our model TS2-Net consistently maintains the state-of-the-art performance on all benchmark datasets with very different characteristics, which demonstrates that our model TS2-Net has decent generalization ability. 

\begin{table*}[]

    \setlength\tabcolsep{4pt} 
	\begin{center}
	\caption{Text-to-Video retrieval results on five benchmarks. We select the previous best performance on each dataset for comparison.}
	\label{table:summary}
        \begin{tabular}{c|c|cccccc} 
            \hline
        	Dataset & Method  & R@1 & R@5 & R@10 & MdR & rsum \\
            \hline
             \multirow{2}{*}{MSR-VTT\cite{xu2016msr}} & CLIP2Video\cite{fang2021clip2video}       & 45.6 & 72.6 & 81.7 & 2.0 & 199.9\\
              & \textbf{TS2-Net(Ours)}             & {\bf 47.0} & {\bf 74.5} & {\bf 83.8} & {\bf 2.0} & {\bf 205.3}\\
             \hline
             
        	 \multirow{2}{*}{VATEX\cite{wang2019vatex}} & CLIP2Video\cite{fang2021clip2video}      & 57.3 & 90.0 & \textbf{95.5} & 1.0 & 242.8\\
              & \textbf{TS2-Net(Ours)}                  & \textbf{59.1} & \textbf{90.0} & 95.2 & \textbf{1.0} & \textbf{244.3}\\
             \hline
             
             \multirow{2}{*}{LSMDC\cite{rohrbach2017movie}} & CAMoE\cite{cheng2021improving}     & 22.5 & \textbf{42.6} & 50.9 & - & 116.0\\
             & \textbf{TS2-Net(Ours)} &  {\bf 23.4}  & 42.3 &  {\bf 50.9} & \textbf{9.0} & \textbf{116.6}\\
             \hline
             
             \multirow{2}{*}{DiDeMo\cite{anne2017localizing}} &             CLIP4Clip\cite{luo2021clip4clip}            & \textbf{42.5} & 70.2 & 80.6 & 2.0 & 193.3\\
             & \textbf{TS2-Net(Ours)}  &       41.8 & {\bf 71.6} &  {\bf 82.0} & \textbf{2.0} & \textbf{195.4}\\
             \hline
             
             \multirow{2}{*}{ActivityNet\cite{caba2015activitynet,krishna2017dense}} & CLIP4Clip\cite{luo2021clip4clip}             & 40.5 & 73.4 & - & 2.0  & -\\
             & \textbf{TS2-Net(Ours)} & {\bf 41.0} & {\bf 73.6} &  {\bf 84.5} & \textbf{2.0} & \textbf{199.1}\\
             \hline
        \end{tabular}
	\end{center}
\end{table*}

\section{More Qualitative Results}
\begin{figure*}[t]
    \centering
    \includegraphics[width=1\linewidth]{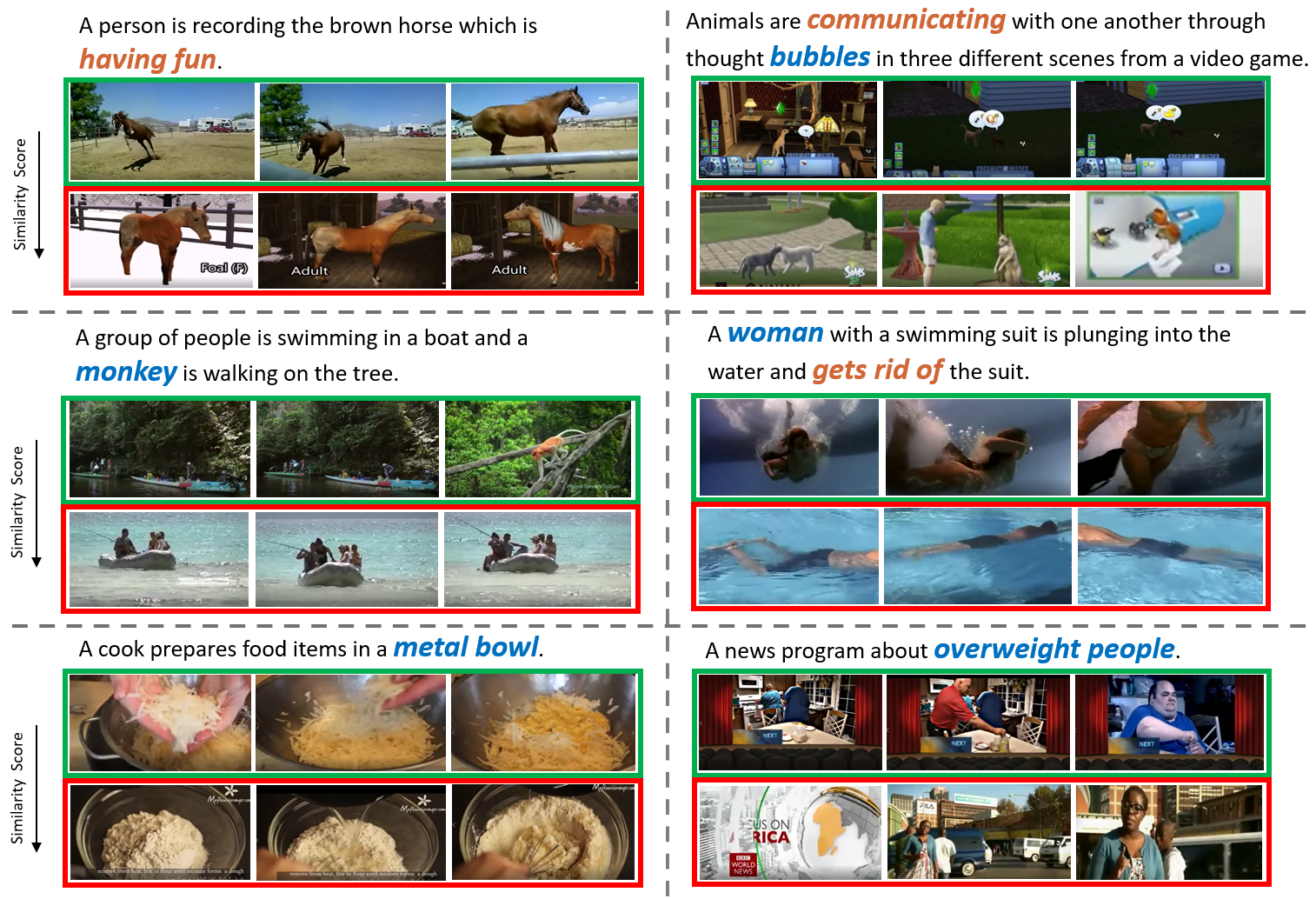}
    \caption{Visualization of more text-video retrieval examples. We rank the retrieval results based on their similarity scores. Green boxed: the correctly retrieved groundtruth video; Red boxed: incorrectly retrieved videos}
    \label{fig:morecases}
\end{figure*}

We provide more qualitative results in Fig.\ref{fig:morecases}. 
In the top left example, our model is able to differentiate the horses that are `having fun' from the horses that are stationary. In the top right example, our model can capture the small object and correctly retrieve the video with `thought bubbles'. Surprisingly, with token shift and token selection module, our model is able to distinguish some \textit{adjective} words. For example, our model correctly retrieves the video with `mental bowl' rather than `glass bowl' (and `overweight people' rather than normal people) in the bottom examples.


\begin{figure*}[ht]
    \centering
    \includegraphics[width=1\linewidth]{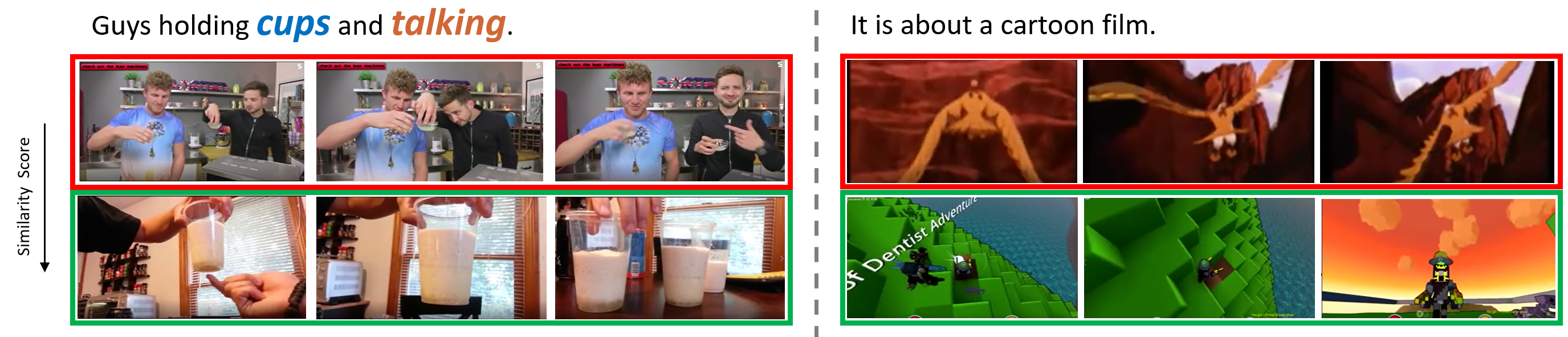}

    \caption{Visualization of some \textit{failure} text-video retrieval examples. We rank the retrieval results based on their similarity scores. Green boxed: the correctly retrieved groundtruth video; Red boxed: the \textit{incorrectly}  retrieved video by our model}
    \label{fig:failurecases}

\end{figure*}

We show some failure cases  as well in Fig.\ref{fig:failurecases}, where our model fails to rank the groundtruth video at the top. 
However, we could argue for these failure cases and consider that our model may actually retrieve the more relevant video. For example, in the left case, the video retrieved by our model (in the red box) seems to be more relevant to the query text, since both `cup' and `talking' can be seen in our results, while the `talking' can not be seen in the ground truth. 

Based on further analysis, we consider that there are also many 
vague and general annotations in the datasets, such as the example shown in the right case in Fig.\ref{fig:failurecases}. Such query annotations account for 1\text{-}2\% of the dataset. We believe our model has a potential to gain in all metrics if such cases get fixed with more discriminative annotations.

\end{document}